\documentclass[]{onethree}

\AtBeginDocument{%
  }

\usepackage{float}
\usepackage{enumitem}
\usepackage{colortbl}
\usepackage{amssymb}
\usepackage{tabularx}
\usepackage{longtable}
\usepackage{needspace}
\providecommand{\Description}[1]{}

\begin{document}

\title{Poplar: A Scalable Pipeline for Human-Centric Image Dataset Synthesis}

\author[1\dagger]{\href{https://choucisan.github.io}{\textcolor{black}{Zhishan Zou}}}
\affiliation[1]{Beijing University of Posts and Telecommunications}



\abstract{
Recent image generators can synthesize convincing human-centric images, yet producing a useful collection remains different from producing a single successful image. A human-centric dataset must cover varied people and contexts, avoid implausible attribute combinations, preserve an everyday photographic character, and expose quality-control decisions at scale.

We present \textbf{Poplar}, a reproducible \emph{Specify--Render--Inspect} pipeline for human-centric image dataset synthesis. \emph{Specify} samples structured attributes under commonsense constraints and verbalizes them as photography-oriented prompts. \emph{Render} uses a realism-adapted image generator across composition-aware aspect ratios and retries obvious technical failures. \emph{Inspect} applies a single structured vision--language review to each candidate, preserving the original prompt while rejecting intrinsic image defects or material prompt mismatches.

Using Poplar, we construct \textbf{Poplar-9K}: 9,401 curated human-centric image--text pairs retained from 11,765 reviewed candidates (79.9\% acceptance). We release the dataset together with the pipeline, configurations, immutable generation prompts, and auditable inspection records as a compact resource for building customizable human-centric collections.
}

% ACM CCS metadata is not used by the OneThree template.

\checkdata[More resources]{
  \href{mailto:choucisan@gmail.com}{\raisebox{-1ex}{\includegraphics[height=4.2ex]{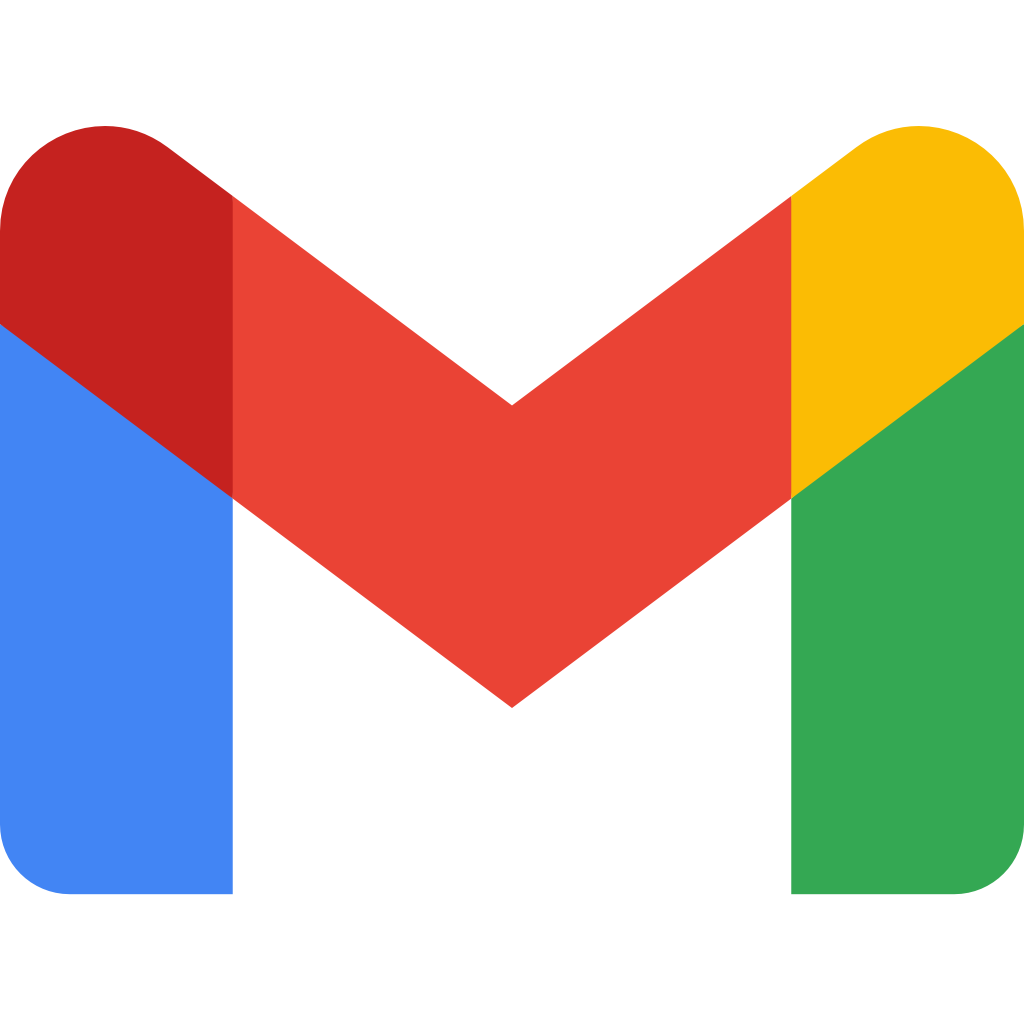}}~Contact}\hspace{1.5em}
  \href{https://choucisan.github.io/publications/poplar}{\raisebox{-1ex}{\includegraphics[height=4.2ex]{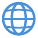}}~Website}\hspace{1.5em}
  \href{https://github.com/choucisan/poplar}{\raisebox{-1ex}{\includegraphics[height=4.2ex]{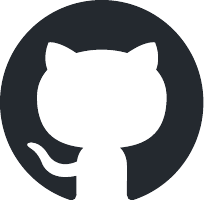}}~Code}\hspace{1.5em}
  \href{https://huggingface.co/collections/choucsan/poplar}{\raisebox{-1ex}{\includegraphics[height=4.2ex]{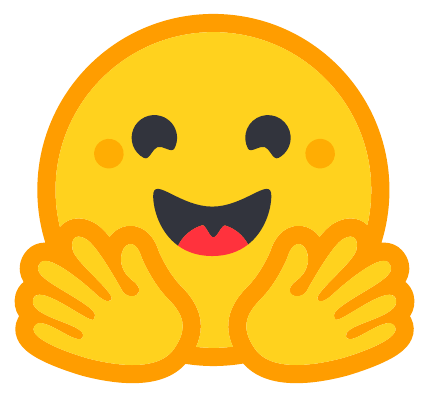}}~Collection}\hspace{1.5em}
  \href{https://www.modelscope.cn/collections/choucisan/Poplar}{\raisebox{-1ex}{\includegraphics[height=4.2ex]{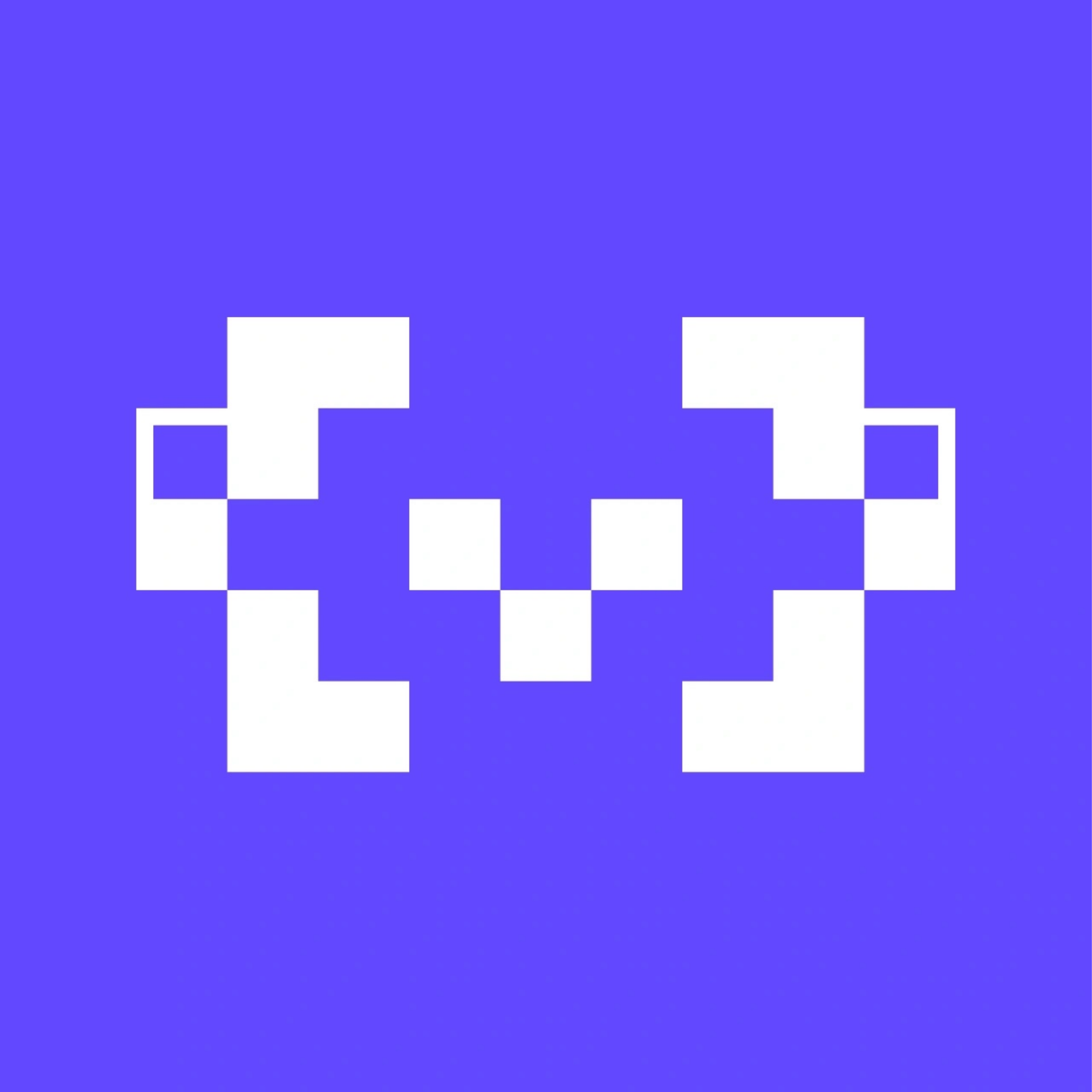}}~Collection}\hspace{1.5em}
}

\vspace*{-6mm}
\maketitle
\vspace{-8mm}

\begin{figure}[h]
  \centering
  \includegraphics[width=1\linewidth]{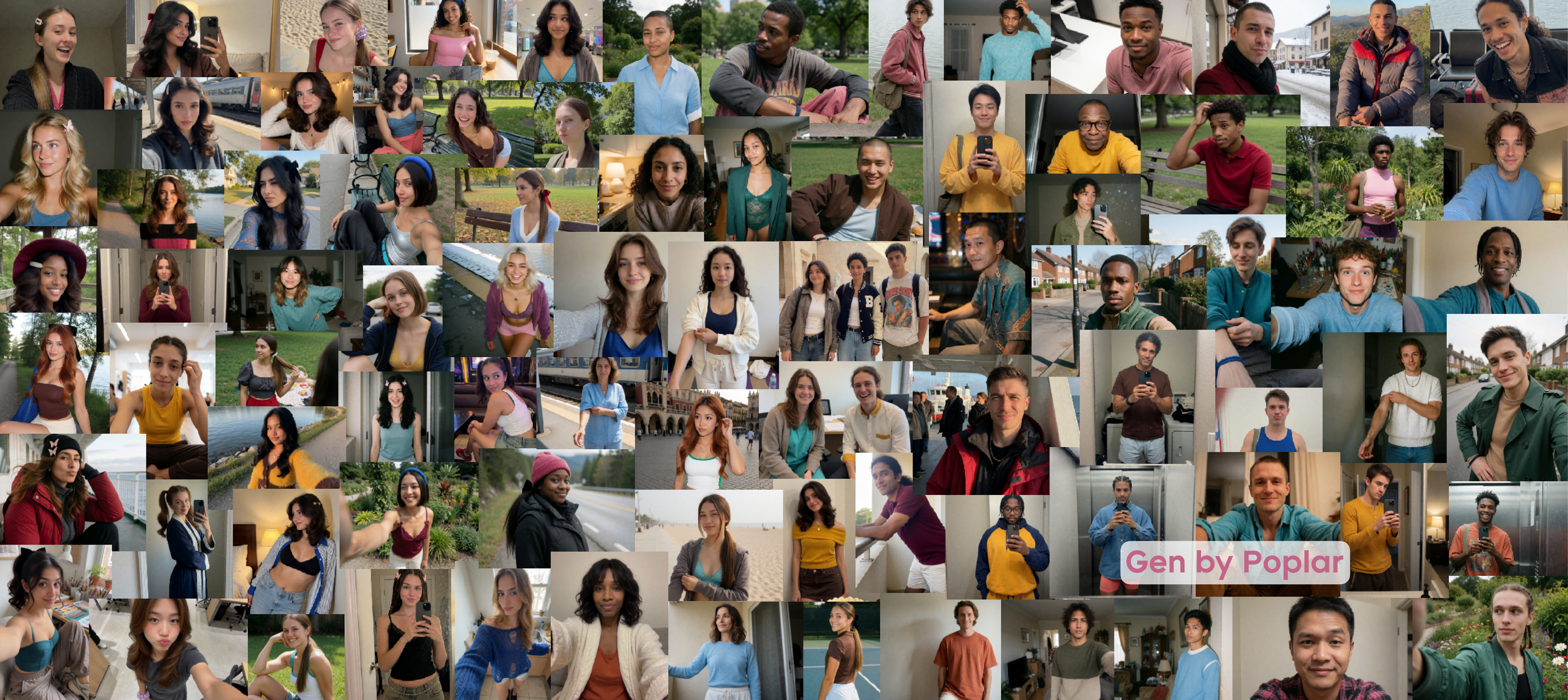}
  \caption{Examples of human-centric images generated by Poplar. The samples
  span diverse subjects, clothing, poses, scenes, camera viewpoints, and
  photographic styles, illustrating the range of everyday human photography
  supported by the pipeline.}
  \label{fig:teaser}
\end{figure}

\section{Introduction}
\label{sec:intro}

Recent advances in diffusion and flow-matching models have substantially improved the fidelity, structural coherence, and text alignment of generated images~\cite{rombach2022ldm,podell2023sdxl,peebles2023dit,lipman2023flow}. Modern systems can now produce individual samples that closely follow complex descriptions and exhibit convincing photographic detail. These advances naturally extend the role of generative models from synthesizing individual images to constructing complete datasets~\cite{tian2023stablerep,hammoud2024synthclip}.

Moving from image synthesis to \emph{dataset synthesis}, however, introduces a different objective. Unlike individual image generation, dataset synthesis seeks to capture not only realistic samples but also the semantic distribution and structural characteristics of a target visual domain. In this work, we focus on \emph{human-centric photography}: images in which people, their activities, interactions, and surrounding contexts constitute the primary visual content. Human-centric photography presents a particularly demanding setting because it combines broad semantic diversity with strong commonsense dependencies among human attributes, activities, social interactions, environments, and photographic conventions. Based on these observations, we identify four desirable properties of synthetic human-centric datasets: \textbf{semantic diversity}, \textbf{realistic attribute co-occurrence}, \textbf{high visual fidelity}, and \textbf{reliable image--text consistency}.

Many existing synthetic data pipelines center on the quality of individual generations or their utility for downstream tasks, rather than the collection-level coverage and structure of everyday human photography. Common workflows expose this gap in several ways: manually authored prompts are expensive to scale and often require iterative refinement~\cite{wang2023diffusiondb,brade2023promptify}; independently sampled attributes increase the number of combinations but can introduce implausible descriptions and distorted co-occurrence patterns; and generation failures or image--text mismatches accumulate across the resulting collection~\cite{huang2023t2icompbench,ghosh2023geneval}. Generative models offer a scalable means to construct such datasets, but only if these collection-level limitations are addressed. This leads to our central question: \emph{How can modern image generators be organized into a reproducible process for constructing diverse and plausible human-centric photographic collections?}

Answering this question requires coordination across the full construction process. Errors introduced during semantic specification propagate through image synthesis, while defects and instruction-following failures accumulate at collection scale. A practical pipeline must therefore connect semantic coverage, commonsense consistency, rendering, and quality control rather than treating them as isolated prompt-by-prompt decisions.

To address these challenges, we present \textbf{Poplar}, a configurable pipeline that unifies three tightly coupled stages for scalable human-centric image dataset synthesis. First, a knowledge-aware prompt producer samples structured attributes under commonsense constraints and converts them into natural photographic descriptions, promoting semantic diversity and realistic attribute co-occurrence. Second, a diffusion-based image producer renders these descriptions as human-centric images with high visual fidelity. Third, a vision--language quality inspector removes samples with visible defects or semantic inconsistencies, improving image--text consistency in the curated collection. Collectively, the three stages connect dataset specification, image synthesis, and quality control to address the four properties identified above.

Using Poplar, we construct \textbf{Poplar-9K}, an open dataset containing 9,401 curated human-centric image--text pairs retained from 11,765 reviewed candidates. The dataset covers diverse subjects, activities, environments, camera viewpoints, and photographic styles. We open-source both Poplar-9K and the Poplar pipeline. Our contributions are deliberately focused and twofold:

\begin{itemize}
    \item We develop and open-source \textbf{Poplar}, a configurable pipeline that unifies knowledge-aware prompt generation, diffusion-based image synthesis, and vision--language quality inspection for scalable human-centric image dataset synthesis.
    \item We construct and release \textbf{Poplar-9K}, an open dataset of 9,401 curated human-centric image--text pairs generated by Poplar, covering diverse subjects, activities, environments, camera viewpoints, and photographic styles.
\end{itemize}

\Needspace{8\baselineskip}
\section{Related Work}
\label{sec:related}

\subsection{Photorealistic Image Generation}

Photorealistic image synthesis has progressed from generative adversarial networks to large text-conditioned diffusion and flow-based models. StyleGAN~\cite{karras2019stylegan} established high-fidelity generation within constrained visual domains. Denoising diffusion probabilistic models~\cite{ho2020ddpm} provided a strong alternative generative formulation, while Imagen~\cite{saharia2022imagen} demonstrated the importance of language understanding for photorealistic text-to-image synthesis. Latent diffusion~\cite{rombach2022ldm} made high-resolution generation more efficient by operating in a compressed representation, and SDXL~\cite{podell2023sdxl} further improved architecture, conditioning, and multi-aspect-ratio training. Diffusion Transformers~\cite{peebles2023dit} and flow matching~\cite{lipman2023flow} subsequently expanded the architectural and training foundations available to modern generators.

Together, these developments provide a mature technical foundation for synthesizing convincing individual images. They do not, however, determine which images should populate a dataset, how semantic factors should co-occur across it, or how generation failures should be controlled at scale.

\subsection{Synthetic Dataset Construction}

Synthetic datasets have traditionally been constructed with procedural engines, simulators, and physically based rendering. Kubric~\cite{greff2022kubric}, for example, provides a scalable generator for controllable scenes with accurate labels. Within human-centric vision, PSP-HDRI+~\cite{ebadi2022psphdri} develops a configurable synthetic-data generator for model pre-training, while Gen4D~\cite{bright2025gen4d} combines an automated 4D human-synthesis pipeline with the SportPAL dataset for sports scenarios. Graphics-based human datasets likewise combine body models, motion capture, scanned assets, and rendering pipelines. Such approaches provide precise control over geometry and annotation, but their coverage is bounded by available assets, motion libraries, scene templates, and rendering configurations.

Foundation models offer a more flexible route: language models can produce semantic specifications, while diffusion models can render them without constructing a complete 3D scene. StableRep~\cite{tian2023stablerep} and SynthCLIP~\cite{hammoud2024synthclip} study synthetic images as supervision for visual representation learning. Real-Fake~\cite{yuan2024realfake} analyzes training-data synthesis from a distribution-matching perspective, and DistDiff~\cite{zhu2024distdiff} expands existing datasets with distribution-consistent samples. DatasetDM~\cite{wu2023datasetdm} and DiffuMask~\cite{wu2023diffumask} further extend diffusion generation to datasets with perception or pixel-level annotations. Together, these studies establish pipeline-based synthesis and distribution-aware construction as active research directions. Their objectives center primarily on downstream representation learning, task-specific perception, or expansion of an existing task dataset. Poplar addresses a complementary setting: constructing open-ended collections of everyday human photography while exposing the semantic specifications, rendering provenance, and quality-control decisions behind the resulting dataset.

\subsection{Human-Centric Image Synthesis}

Human-centric image synthesis has evolved from graphics-based control to flexible generation with large pretrained models. SURREAL~\cite{varol2017surreal} renders motion-capture sequences with 3D body models to obtain large-scale human supervision; AGORA~\cite{patel2021agora} and BEDLAM~\cite{black2023bedlam} increase realism, clothing variation, multi-person composition, and body annotation quality. These resources provide strong geometric control but remain tied to body assets, motion libraries, and rendering environments. Diffusion models broadened the setting to human image generation and editing from natural language. DreamBooth~\cite{ruiz2023dreambooth} supports subject-driven personalization, while ControlNet~\cite{zhang2023controlnet} conditions generation on pose, layout, depth, and other structural cues. This line of work has substantially improved the fidelity and controllability of individual human images.

Everyday human photography, however, is not defined by a single person, pose, or editing target. A collection contains long-tail variation and dependencies among subjects, activities, clothing, social interactions, environments, viewpoints, compositions, and photographic styles. Producing a requested image and constructing a dataset with this structure are distinct objectives: the latter additionally requires semantic coverage, plausible co-occurrence, and consistent quality across many samples. Poplar approaches human-centric photography from this dataset-level perspective, connecting semantic specification, image generation, and quality inspection within one configurable pipeline.

\section{Poplar}
\label{sec:poplar}

\begin{figure}[H]
  \centering
  \includegraphics[width=\linewidth]{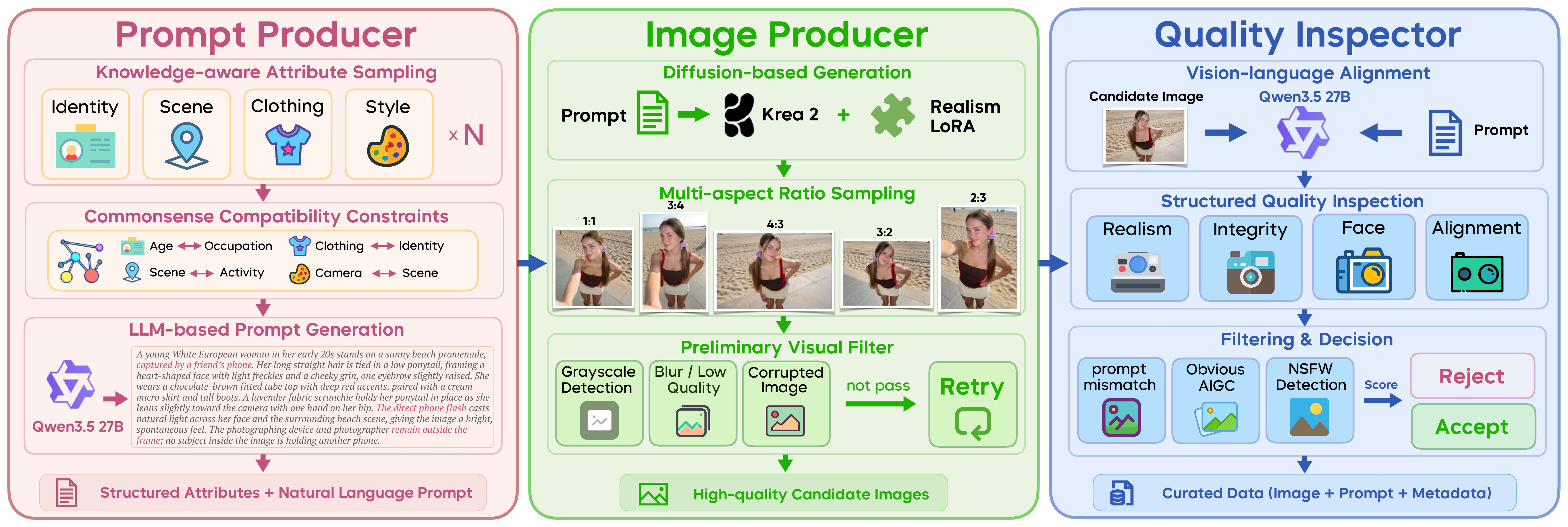}
  \caption{Overview of Poplar. A knowledge-aware Prompt Producer constructs
  structured attributes and natural-language prompts; a diffusion-based Image
  Producer renders and visually pre-filters candidates; and a vision--language
  Quality Inspector verifies semantic alignment and returns curated images,
  prompts, and metadata.}
  \label{fig:poplar_pipeline}
\end{figure}

Poplar converts a dataset specification into curated human-centric image--text pairs through three explicit contracts. \textbf{Specify} produces an immutable natural-language prompt, a structured semantic record, a composition-aware canvas, and a rendering seed. \textbf{Render} consumes this record and returns a decodable candidate plus a manifest of model and pre-filter settings. \textbf{Inspect} consumes exactly one image--prompt pair and returns a structured keep/reject decision with visible evidence. As shown in Figure~\ref{fig:poplar_pipeline}, the corresponding Prompt Producer, Image Producer, and Quality Inspector remain replaceable, while JSONL records make every transition resumable and auditable.

\subsection{Knowledge-aware Prompt Producer}
\label{sec:prompt_producer}

The semantic coverage and internal coherence of a synthetic dataset are shaped before image generation begins. Directly asking an LLM to produce free-form prompts entangles two different decisions: \emph{what} the dataset should contain and \emph{how} each sample should be described. Repeating this process at scale can overrepresent concepts preferred by the LLM, leave important regions of the intended attribute space uncovered, and make collection-level composition difficult to inspect or adjust. Poplar therefore separates content planning from linguistic realization. Its Prompt Producer comprises three steps: photography-oriented prompt design, knowledge-aware attribute sampling, and LLM-based prompt generation.

\subsubsection{Photography-oriented Prompt Design}
\label{sec:photography_prompt}

Modern image generators can render a described person with high visual fidelity, yet a content-only prompt often defaults to a polished or staged composition. Everyday human photographs follow a different capture process. They include casual snapshots, selfies, and photos taken by friends or family, with ordinary viewpoints, direct phone flash, uneven natural illumination, loose framing, and other non-studio characteristics. Our goal is not to deliberately degrade image quality, but to describe the conditions under which an ordinary photograph would plausibly be captured.

We therefore organize each prompt into two complementary layers. The \emph{semantic layer} specifies the subject, clothing, activity, and surrounding scene. The \emph{capture layer} describes the photographer's perspective, camera or phone characteristics, viewpoint, framing, lighting, and plausible photographic imperfections. It also resolves viewpoint-specific ambiguities, for example by stating whether the device and photographer remain outside the frame or whether a visible phone is held by the subject. This capture-oriented language encourages the generator to realize the semantic specification as an everyday photograph rather than an idealized portrait.

\begin{figure}[t]
  \centering
  \includegraphics[width=\linewidth]{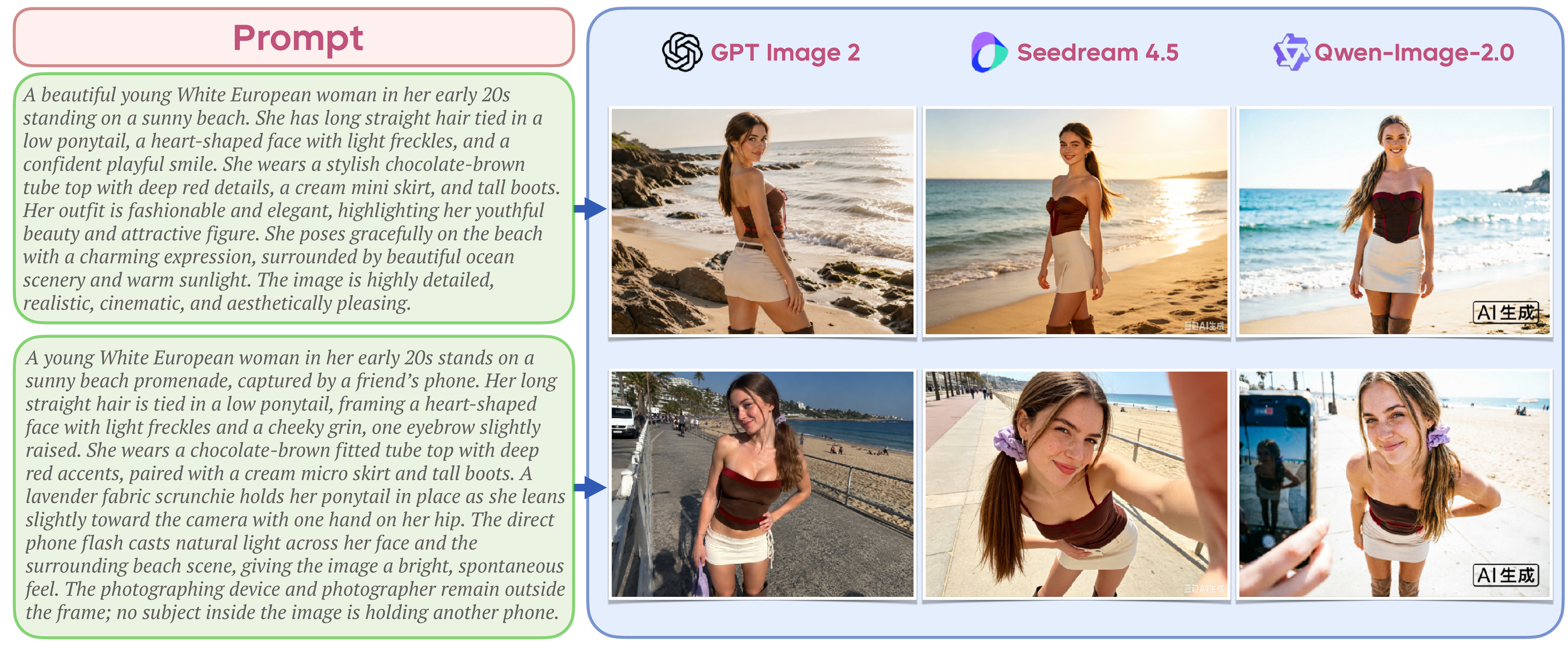}
  \caption{Qualitative comparison of content-centric (top) and
  photography-oriented (bottom) prompts across three image generators. Adding
  capture conditions, photographer perspective, and framing constraints shifts
  the outputs from staged portraits toward casual photographic compositions.}
  \label{fig:prompt_design}
\end{figure}

Figure~\ref{fig:prompt_design} qualitatively illustrates this distinction. With broadly similar subject and scene content, a conventional prompt produces clean, carefully composed portraits across several generators, whereas the photography-oriented version introduces an explicit photographer, capture device, direct flash, and framing constraints, leading to more casual viewpoints and compositions.

\subsubsection{Knowledge-aware Attribute Sampling}
\label{sec:attribute_sampling}

Free-form LLM generation provides little direct control over which concepts appear or how frequently they occur. To make dataset composition explicit, Poplar first represents each candidate with a structured attribute configuration
\begin{equation}
  \mathbf{a}=(a_1,a_2,\ldots,a_m), \qquad a_i\in\mathcal{A}_i,
  \label{eq:attribute_configuration}
\end{equation}
where each \(\mathcal{A}_i\) is a configurable attribute space. The current design covers subject and identity descriptors, clothing, pose or action, scene, camera viewpoint, lighting, composition, and photographic style. Each dimension is sampled from an explicit proposal distribution, allowing a dataset builder to inspect and adjust its marginal coverage rather than inheriting the implicit preferences of an LLM.

Independent sampling alone is insufficient because valid values can form an invalid combination. Poplar therefore evaluates sampled configurations against a collection of commonsense compatibility constraints,
\begin{equation}
  C(\mathbf{a})=\bigwedge_{k=1}^{K}c_k(\mathbf{a}),
  \label{eq:commonsense_constraints}
\end{equation}
and retains or repairs a configuration only when the relevant constraints are satisfied. The implemented rules cover age--scene and age--clothing feasibility, gender--clothing and cultural-background--clothing compatibility, clothing--scene consistency, and agreement among subject count, camera perspective, and photographic context. Higher-order rules assign distinct faces, hairstyles, expressions, poses, colors, and outfits in multi-person scenes. Importantly, these constraints target physical or semantic contradictions rather than enforcing a single aesthetic preference or removing uncommon but plausible combinations. The result is an inspectable attribute pool with fewer internally inconsistent specifications.

\subsubsection{LLM-based Prompt Generation}
\label{sec:llm_prompt_generation}

After validation, Poplar provides the structured configuration, the photography-oriented template, and a set of generation rules to an LLM. In our implementation, Qwen3.5-27B-FP8~\cite{qwen2026qwen35} acts as a verbalizer rather than an unconstrained content generator. It connects the sampled fields into a coherent natural-language description, adds only the contextual details needed for fluency, and preserves the selected subject, activity, clothing, environment, and capture conditions. For multi-person scenes, the instruction assigns attributes and actions unambiguously to each subject and prevents details from being exchanged across people. Canvas selection remains outside the LLM: the model returns only an identifier and prompt, preventing linguistic realization from silently changing the planned composition.

\begin{figure}[H]
  \centering
  \includegraphics[width=\linewidth]{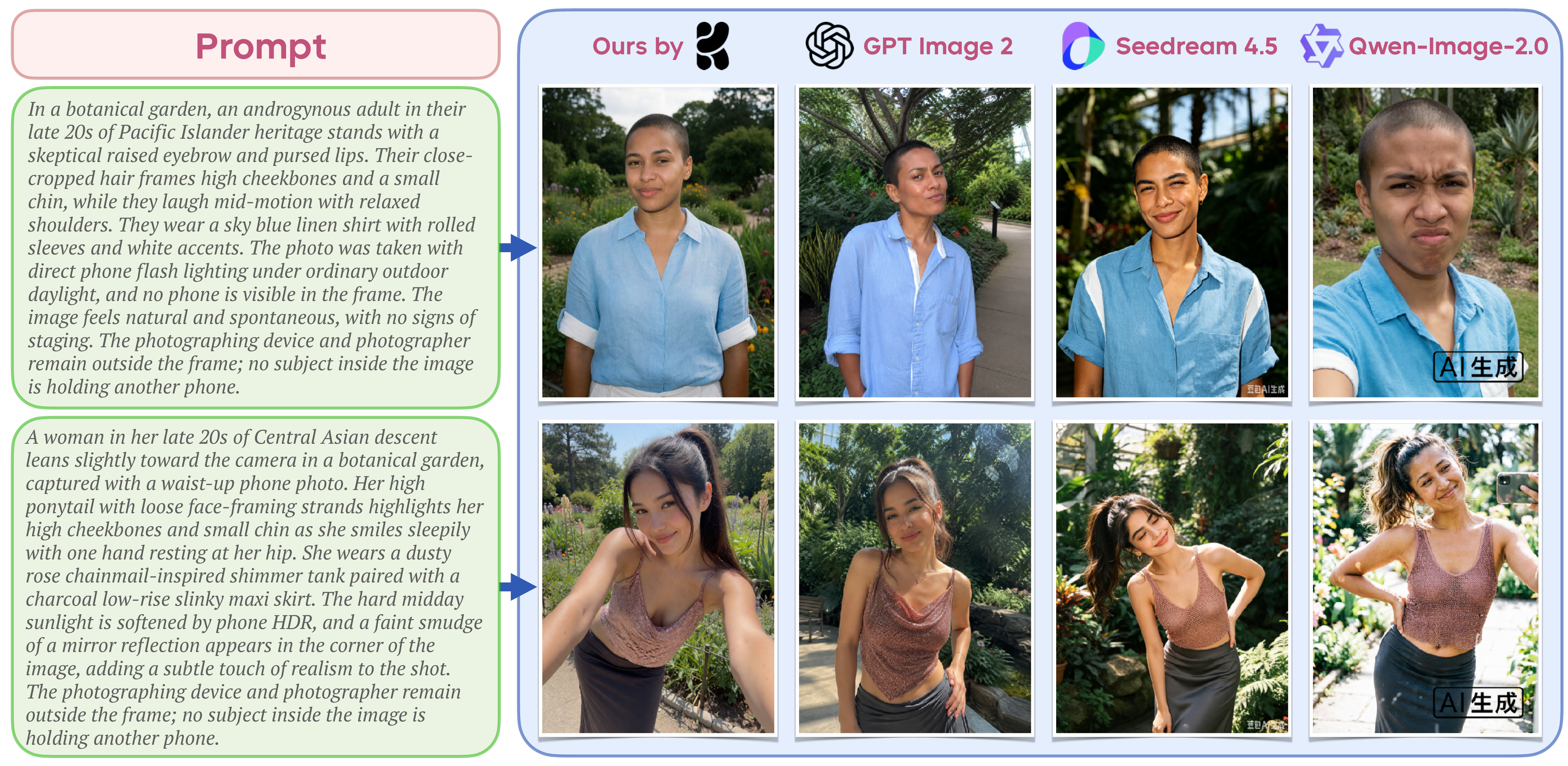}
  \caption{Examples produced from Poplar prompts using our Image Producer and
  three external image generators. Although rendering characteristics vary,
  the examples illustrate that the same structured semantic content and
  photography-oriented capture conditions can be used with different generator
  families.}
  \label{fig:prompt_examples}
\end{figure}

Formally, the prompt is generated as
\begin{equation}
  p=f_{\mathrm{LLM}}(\mathbf{a},t_{\mathrm{photo}},r),
  \label{eq:prompt_generation}
\end{equation}
where \(t_{\mathrm{photo}}\) denotes the photography-oriented template and \(r\) contains consistency and formatting rules. Poplar stores \(p\) together with \(\mathbf{a}\), preserving both a natural-language condition for image generation and structured metadata for later inspection. Because semantic selection is separated from verbalization, the same prompt specification can be rendered by different image generators, as illustrated in Figure~\ref{fig:prompt_examples}.

\subsection{Diffusion-based Image Producer}
\label{sec:image_producer}

Given a photography-oriented prompt \(p\) and its structured attributes \(\mathbf{a}\), the Image Producer renders candidate images at scale. Poplar does not introduce a new generative model; instead, it treats a pretrained image generator as a replaceable rendering backend. This design separates dataset construction from a particular model architecture and allows improvements in image generation to be incorporated without changing the upstream Prompt Producer or downstream Quality Inspector.

\noindent\textbf{Rendering backbone.}
In our implementation, we use Krea~2 Turbo~\cite{krea2026krea2} with the Krea2-realism-V2 adapter~\cite{rudysen2026krea2realism}. The prompt and adapter serve complementary roles: the prompt specifies the subject, scene, and photographic capture process, while the adapter biases rendering toward natural textures, lighting, and camera appearance. This backbone is an implementation choice rather than a requirement of Poplar; as illustrated in Figure~\ref{fig:prompt_examples}, the same prompts can also condition other image generators.

\noindent\textbf{Multi-aspect-ratio generation.}
A dataset rendered on a single canvas inherits a narrow compositional prior. Poplar samples one target aspect ratio for each specification from
\begin{equation}
  \rho\sim q_{\rho}(\,\cdot\mid\mathbf{a}), \qquad
  \rho\in\mathcal{R}=\{1{:}1,\,2{:}3,\,3{:}2,\,3{:}4,\,4{:}3\},
  \label{eq:aspect_ratio_sampling}
\end{equation}
where the configurable distribution is conditioned on composition-relevant attributes in \(\mathbf{a}\). Mirror selfies and full-body photographs favor portrait canvases, whereas multi-person and environmental photographs favor moderate landscape canvases. The selected ratio and dimensions are stored before verbalization and rendering. Thus, the five formats diversify the collection; they do not imply that every prompt is rendered five times.

\noindent\textbf{Preliminary visual filtering.}
Even strong generators occasionally produce obvious technical failures during large-scale inference. Before invoking the more expensive vision--language inspection, Poplar applies a deterministic image-only filter. It measures mean RGB chroma and the fraction of nearly neutral pixels to identify near-grayscale outputs, and computes the variance of a grayscale edge response to detect severe blur or abnormally low detail. Each saved PNG is then reopened and verified to catch truncation or decoding failure. Thresholds are deliberately conservative: this stage removes technical failures rather than assigning an aesthetic score. A failed candidate is archived with its metrics and returned to the rendering queue with the same prompt and a new seed. The final manifest records the requested seed, actual seed, retry count, accepted metrics, and every failed attempt.

\subsection{Vision-Language Quality Inspector}
\label{sec:quality_inspector}

Passing the preliminary visual filter does not necessarily make a generated image a suitable training sample. During large-scale generation, modern image generators may only partially follow an instruction, omit important human attributes, or produce compositions that appear plausible at first glance but contain duplicated subjects, stitched regions, abnormal person layouts, or insufficiently visible faces. They may also generate nudity or other sensitive content. If retained, such samples introduce noisy supervision and distort the semantic and visual characteristics of the resulting dataset. Poplar therefore introduces a Vision--Language Quality Inspector as the final gate of the synthesis pipeline.

\begin{figure}[H]
  \centering
  \includegraphics[width=\linewidth]{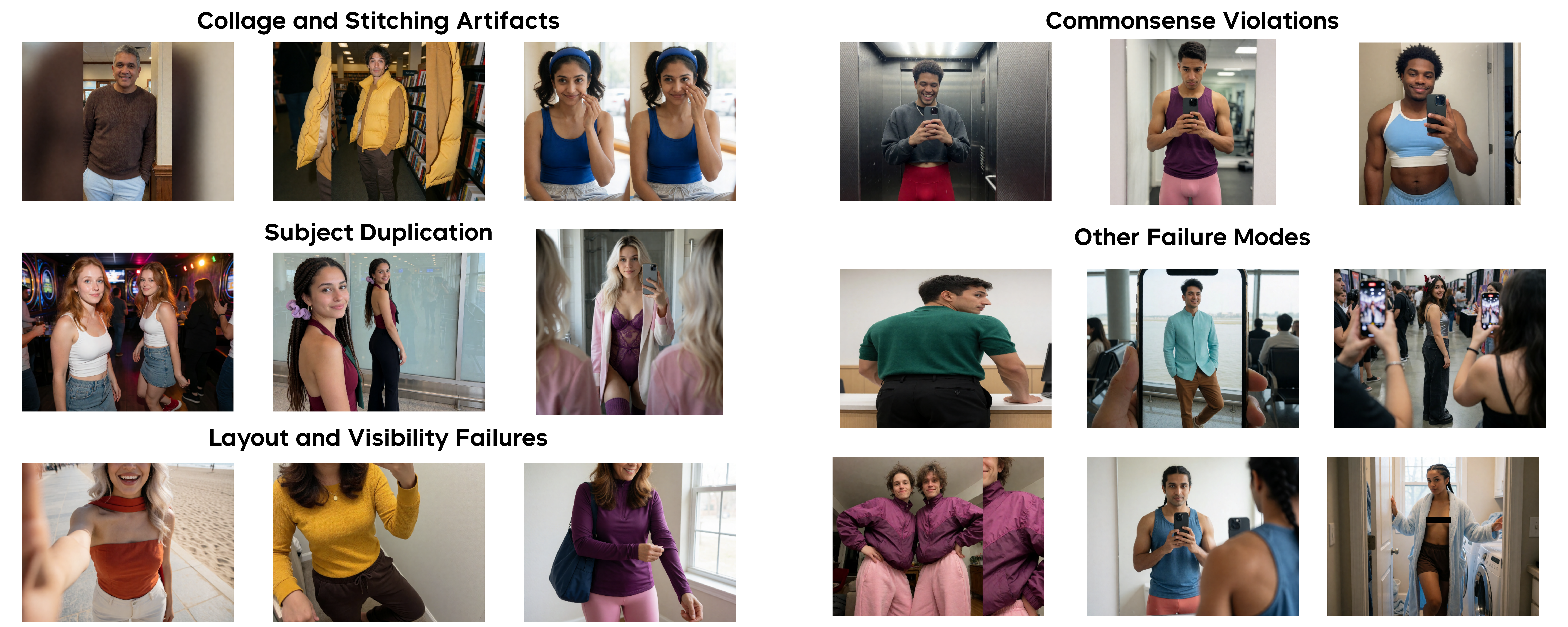}
  \caption{Representative samples rejected by the Quality Inspector. The examples exhibit collage and stitching artifacts, subject duplication, commonsense violations, invalid human layouts or insufficient subject visibility, and other generation failure modes. Sensitive regions are redacted for presentation.}
  \label{fig:rejected_samples}
\end{figure}

Figure~\ref{fig:rejected_samples} shows representative candidates rejected during this stage. These examples illustrate that many failures are not captured by file-level checks alone: an image can be decodable and visually sharp while still violating the semantic, compositional, or safety requirements of the dataset.

\noindent\textbf{Structured quality criteria.}
Given a generated image \(x\) and its immutable prompt \(p\), the Inspector first checks intrinsic image quality and then prompt consistency. Its intrinsic defect flags cover obvious synthetic appearance, collage or duplicated layout, missing or cropped faces, anatomy or object-interaction errors, implausible content, explicit nudity, and monochrome or corrupt output. A separate mismatch list compares subject count, gender presentation, scene, clothing, age range, and face presence against the prompt, with critical, major, or minor severity. The output additionally records visible people, clothing descriptions, face visibility, confidence, and concise evidence. These fields preserve the reason for every rejection instead of reducing review to an opaque score.

\noindent\textbf{Vision-language inspection.}
TIFA~\cite{hu2023tifa} uses question answering for fine-grained and interpretable text-to-image faithfulness evaluation, while LLaVA-Critic~\cite{xiong2024llavacritic} demonstrates the broader use of multimodal models as evaluators. Building on this direction, Poplar uses Qwen3.5-27B-FP8 once per image to apply a fixed inspection rubric and return a structured decision:
\begin{equation}
  \mathcal{V}(x,p)
  \rightarrow
  \bigl(\mathbf{d},\mathbf{m},\mathbf{v},e\bigr),
  \label{eq:vlm_inspection}
\end{equation}
where \(\mathbf{d}\) contains binary intrinsic defect flags, \(\mathbf{m}\) contains prompt mismatches and their severities, \(\mathbf{v}\) describes visible people, and \(e\) records concise evidence. A sample is admitted only when no intrinsic defect is present and no critical or major mismatch is reported:
\begin{equation}
  b_{\mathrm{accept}}
  = \neg\!\left(\bigvee_i d_i\right)
    \land \neg\!\left(\bigvee_j [s(m_j)\in\{\mathrm{critical},\mathrm{major}\}]\right).
  \label{eq:quality_acceptance}
\end{equation}
Using structured criteria makes the filtering process configurable and auditable, while avoiding an ambiguous single overall quality score.

\noindent\textbf{Filtering and feedback.}
Rejected images remain associated with their failure codes for audit but do not enter the release. Images that pass retain the original generation prompt byte-for-byte: the Inspector is not allowed to caption, paraphrase, or repair it. This distinction matters because prompt construction is a primary artifact of Poplar; changing prompts after viewing the output would hide instruction-following errors and weaken reproducibility.

\section{Poplar-9K}
\label{sec:poplar9k}

To demonstrate the complete Poplar workflow in practice, we use the pipeline to construct \textbf{Poplar-9K}, a curated dataset of everyday human-centric photography. Poplar-9K is not intended merely as a collection of successful generations; it records the output of a dataset-level construction process in which semantic configurations are sampled explicitly, rendered at scale, and subjected to both technical and semantic quality control.

\subsection{Dataset Construction}

We deploy Poplar on four NVIDIA GeForce RTX~4090 GPUs. The construction run yields 11,765 candidates submitted to the Quality Inspector. Each candidate is paired with its photography-oriented prompt and structured attributes and rendered at one of the five composition-aware aspect ratios. The preliminary filter retries grayscale, severely blurred, or invalid files before review. The Inspector then removes prompt--image mismatches, synthetic-looking outputs, collage and duplication artifacts, implausible human--scene relationships, incomplete faces, anatomy or object errors, and sensitive content.

After curation, 9,401 images remain and 2,364 are rejected, giving an acceptance rate of 79.9\%. The resulting collection pairs every retained image with its unchanged generation prompt and structured semantic attributes. This fixed funnel---11,765 reviewed, 9,401 retained, and 2,364 rejected---is the complete construction account reported for Poplar-9K.

\subsection{Dataset Statistics}

\begin{figure}[H]
  \centering
  \includegraphics[width=\linewidth]{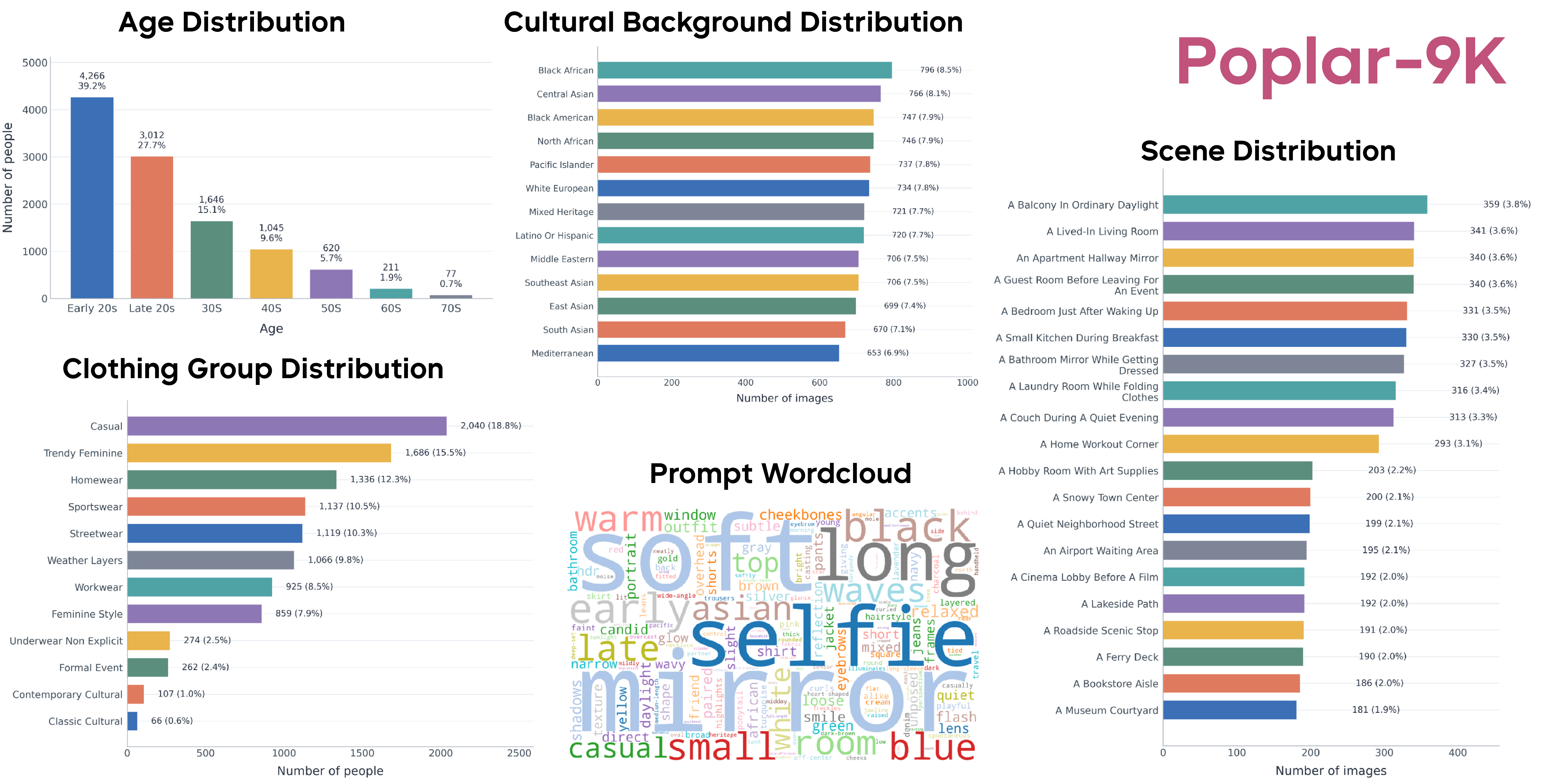}
  \caption{Statistics of Poplar-9K. The dataset contains 9,401 curated images and 10,877 depicted people. The panels summarize age, cultural background, clothing group, frequent prompt terms, and representative scene frequencies.}
  \label{fig:poplar9k_stats}
\end{figure}

Figure~\ref{fig:poplar9k_stats} summarizes the composition of Poplar-9K. The 9,401 images contain 10,877 depicted people because a subset of the collection contains multiple subjects. Younger adults form the largest age groups, while the dataset also includes subjects from their 30s through 70s. Across 13 sampled cultural-background descriptors, individual groups account for approximately 6.9\%--8.5\% of the images, indicating relatively balanced coverage under the adopted sampling configuration. These statistics are computed from the structured prompt attributes; the labels describe the synthesis specification rather than identities inferred from the generated images.

The clothing distribution spans casual clothing, contemporary fashion, homewear, sportswear, streetwear, workwear, formal attire, and culturally specific styles. Casual clothing is the most frequent group at 18.8\%, followed by trendy feminine styles at 15.5\% and homewear at 12.3\%. Scene frequencies exhibit a broader long tail: even the most frequent listed setting accounts for only 3.8\% of the collection. The scenes cover domestic interiors, neighborhood streets, public venues, travel settings, and outdoor environments, while the prompt word cloud provides a complementary view of the descriptive vocabulary used during synthesis.

\subsection{Open-source Release}

We open-source two complementary artifacts: the \textbf{Poplar-9K dataset} and the \textbf{complete Poplar codebase}. The dataset release contains the 9,401 curated images together with their generation prompts and structured attributes, making its composition inspectable and allowing the semantic specifications to be reused with other image generators. The code release implements the Prompt Producer, Image Producer, and Quality Inspector, and exposes the attribute space, sampling distributions, generation backend, and inspection policy as configurable components. Poplar-9K represents one public instantiation rather than the scale limit of the pipeline: by increasing the generation budget and expanding the attribute and scene spaces, the same workflow can be used to construct larger and more varied image collections. Together, the dataset and code provide both a concrete output of Poplar and a reproducible implementation that researchers can adapt to other target domains.

\subsection{Reproducibility Contract}

Datasheets for Datasets~\cite{gebru2021datasheets} and Data Cards~\cite{pushkarna2022datacards} emphasize communicating a dataset's motivation, composition, construction process, provenance, and intended use to downstream users. Motivated by this principle, Poplar couples its human-readable dataset description with an executable construction record. The released configuration is the executable description of the Poplar-9K construction policy. It fixes the taxonomy version, random seeds, candidate budget, model identifiers, aspect-ratio set, rendering parameters, pre-filter thresholds, and inspection rubric. Prompt specifications are generated in two resumable shards with Qwen3.5-27B-FP8. Rendering uses Krea~2 Turbo with the Krea2-realism-V2 adapter at scale 1.5, eight inference steps, zero guidance, and canvases whose longer side does not exceed 1,024 pixels. Inspection uses Qwen3.5-27B-FP8 with deterministic decoding and one model invocation per image. Every stage appends JSONL records and skips completed identifiers, so an interrupted run resumes without silently replacing earlier samples.

For each retained item, the release preserves four layers of provenance: the immutable prompt and its hash; the structured attributes and taxonomy version; rendering settings, dimensions, requested and actual seeds, and preliminary-filter history; and the final structured inspection record. Rejected candidates retain their issue category, severity, visible evidence, and confidence in a separate audit manifest. This design makes the 11,765-to-9,401 curation funnel inspectable rather than presenting Poplar-9K as an unexplained collection of successful generations.

\FloatBarrier

\section{Scope and Limitations}

Poplar is a dataset-construction project, not a new image generator, language model, or learned quality metric. Poplar-9K is intentionally modest in scale and demonstrates one configurable sampling policy rather than claiming to reproduce the full distribution of human photography. Its age and cultural-background statistics describe requested synthesis attributes; they must not be interpreted as verified demographic labels or inferred identities. The compatibility rules reduce obvious contradictions but inevitably encode design choices, and uncommon yet plausible combinations may be underrepresented.

Automated inspection also remains fallible. A vision--language model can miss subtle artifacts or reject an acceptable image, and deterministic pixel checks cover only narrow technical failures. We therefore release inspection evidence and rejected-item metadata so that users can audit or replace the policy. Finally, generated people are synthetic and should not be treated as records of real individuals. Applications involving identity, demographic measurement, or high-stakes decisions require validation beyond the scope of this project.

\section{Conclusion}
\label{sec:conclusion}

In this work, we study human-centric image dataset synthesis as a collection-level problem rather than an extension of isolated image generation. Our central question is how to organize modern generators into a reproducible process for constructing diverse and plausible everyday photographic collections. From this perspective, we treat semantic diversity, realistic attribute co-occurrence, visual fidelity, and image--text consistency as joint design goals across dataset construction.

To address these requirements, we present Poplar, a configurable pipeline that connects knowledge-aware semantic specification, diffusion-based image generation, and vision--language quality inspection. The Prompt Producer controls dataset composition through structured attribute sampling and photography-oriented descriptions; the Image Producer renders these specifications across multiple photographic formats; and the Quality Inspector rejects technically valid but semantically inconsistent, compositionally defective, or unsafe outputs. Poplar is designed around replaceable components, allowing its attribute space, generation backend, and inspection policy to evolve independently.

Using Poplar on four NVIDIA GeForce RTX~4090 GPUs, we review 11,765 candidate images and retain 9,401 after curation, forming the open-source Poplar-9K dataset. Poplar-9K is a concrete, small-scale instantiation of the construction process, released with its prompts, structured attributes, rendering provenance, and inspection records. The contribution is practical by design: a reusable way to grow human-centric image collections and a compact dataset that exposes what this process produces, rejects, and records.

\clearpage
\bibliographystyle{plainnat}
\bibliography{poplar}

\end{document}